\documentclass{article}

    \PassOptionsToPackage{numbers, compress}{natbib}

    \usepackage[preprint]{neurips_2021}

\usepackage{amsmath,amsfonts,bm}

\newcommand{\sect}[1]{$\mathsection$#1}

\newcommand{\figleft}{{\em (left)}}
\newcommand{\figcenter}{{\em (centre)}}
\newcommand{\figright}{{\em (right)}}

\def\eqref#1{equation~\ref{#1}}

\def\1{\bm{1}}

\def\rx{{\textnormal{x}}}
\def\ry{{\textnormal{y}}}

\def\ve{{\bm{e}}}

\def\vy{{\bm{y}}}
\def\vz{{\bm{z}}}

\def\mX{{\bm{X}}}
\def\mY{{\bm{Y}}}

\DeclareMathAlphabet{\mathsfit}{\encodingdefault}{\sfdefault}{m}{sl}
\SetMathAlphabet{\mathsfit}{bold}{\encodingdefault}{\sfdefault}{bx}{n}

\def\sF{{\mathbb{F}}}

\def\sR{{\mathbb{R}}}

\def\sV{{\mathbb{V}}}

\newcommand{\sigmoid}{\sigma}

\DeclareMathOperator*{\argmax}{arg\,max}

\usepackage[utf8]{inputenc} 
\usepackage[T1]{fontenc}    
\usepackage{hyperref}       
\usepackage{url}            
\usepackage{booktabs}       
\usepackage{amsfonts}       
\usepackage{nicefrac}       
\usepackage{microtype}      

\usepackage{upgreek}

\usepackage{color}
\usepackage{xcolor} 
\usepackage{colortbl}
\usepackage{collcell} 

\usepackage{amsmath}
\usepackage{amssymb}
\usepackage{relsize}        
\usepackage{tikz}
\usepackage{tkz-euclide}
\usepackage{float}
\usepackage{todonotes}
\usepackage{subcaption}
\usepackage{bbm}
\usepackage{bm}
\usepackage{enumitem}       
\usepackage[utf8]{inputenc}
\usepackage{multirow}
\usepackage{rotating}
\usepackage{cleveref}
\usepackage{diagbox}
\usepackage[title]{appendix}

\usetikzlibrary{bayesnet}
\usepackage{url}
\usepackage{soul}       
\usepackage{dblfloatfix}    

\title{A Probabilistic Model for Discriminative and Neuro-Symbolic Semi-Supervised Learning}

\author{%
  \hspace{.3cm} Carl Allen      \hspace{1.8cm}       Ivana Bala\v{z}evi\'c   \hspace{.8cm}       Timothy Hospedales \\
  University of Edinburgh     \qquad Samsung AI Centre, Cambridge \\
  \texttt{\{carl.allen,   ivana.balazevic,    t.hospedales\}@ed.ac.uk}
}

\begin{document}

\maketitle

\begin{abstract}
    Much progress has been made in semi-supervised learning (SSL) by combining methods that exploit different aspects of the data distribution, e.g.~\textit{consistency regularisation} relies on properties of $p(x)$, whereas \textit{entropy minimisation} pertains to the label distribution $p(y|x)$.
    Focusing on the latter, we present a probabilistic model for  \textit{discriminative} SSL, that mirrors its classical generative counterpart. 
    Under the assumption $y|x$ is \textit{deterministic}, the prior over latent variables becomes \textit{discrete}. 
    We show that several well-known SSL methods can be interpreted as approximating this prior, and can be improved upon.
    We extend the discriminative model to \textit{neuro-symbolic} SSL, where label features satisfy logical rules, by showing such rules relate directly to the above prior, thus justifying a family of methods that link statistical learning and logical reasoning,
    and unifying them with regular SSL.
\end{abstract}

\vspace{-5pt}
\section{Introduction}
\label{sec:intro}
\vspace{-5pt}

Semi-supervised learning (SSL) learns to predict a label $y$ for each data point $x$ from labelled data $\{(x^i,y^i)\}$ and a set of, often more abundant, unlabelled data $\{x^j\}$. 
For unlabelled data to help predict labels, the distribution $p(x)$
must contain information relevant to that prediction \cite{chapelle2009semi, zhu2009introduction}. 
State-of-the-art SSL algorithms \citep[e.g.][]{berthelot2019mixmatch, berthelot2019remixmatch} combine several underlying methods, some of which directly leverage properties of $p(x)$, such as \textit{data augmentation} and \textit{consistency regularisation} \citep{sajjadi2016regularization, laine2017temporal,tarvainen2017mean, miyato2018virtual}. 
Others utilise properties of the conditional label distributions $p(y|x)$ by adding a bespoke function of the model's predictions for unlabelled data to a standard supervised loss function, e.g.~\textit{entropy minimisation} \citep{grandvalet2005semi}, \textit{mutual exclusivity} \citep{sajjadi2016mutual, xu2018semantic} and \textit{pseudo-labelling} \citep{lee2013pseudo}. We refer to such methods as \textit{discriminative} semi-supervised learning (DSSL) and show that they can be justified and unified under a probabilistic model, comparable to the classical generative model for SSL \cite{chapelle2009semi, zhu2009introduction, van2020survey}.

In some tasks, vector labels indicate the presence/absence of a set of attributes that obey logical rules, e.g.~\textit{legs}$\,\Rightarrow\! \neg$\,\textit{fins}. A neural network-based SSL algorithm that takes such rules into account combines statistical machine learning with logical reasoning, a paradigm known as \textit{neuro-symbolic learning} (NSL). 
Several methods for neuro-symbolic SSL \cite[e.g.][]{xu2018semantic, van2019semi} add a term based on logical constraints to a supervised loss function. We show that such methods, although often disjoint from `regular' SSL in the literature, are also theoretically justified under the proposed probabilistic model for discriminative SSL.
Thus, within the scope considered, the DSSL model provides a principled basis for integrating logical reasoning and statistical learning.

The proposed DSSL model is a hierarchical latent variable model in which each data point $x\!\in\!\mathcal{X}$ has an associated label distribution $p(y|x)$ with parameter $\theta\!\in\!\Theta$. Parameters $\theta$ are treated as latent random variables sampled from a distribution $p(\theta)$.
A parametric function $f_\omega\!:\mathcal{X}\!\to\!\Theta$ (e.g.~a neural network) is assumed to learn $\theta$ as a function of $x$, $\smash{f_\omega(x)\!\doteq\!\tilde\theta \approx \theta}$; e.g.~in $K$-class classification, $f_\omega(x)$ maps to a particular multinomial parameter on the simplex $\Delta^K\!\!\subset\!\sR^K\!$.\footnote{
    While we focus on classification as a common SSL use-case, the DSSL model generalises to other tasks.} 
It follows that the empirical distribution of model outputs $\smash{\tilde\theta}$ is expected to follow $p(\theta)$. In particular, the distribution of outputs for \textit{unlabelled} data should accord with $p(\theta)$, and $f_\omega$ can be updated if not -- providing a learning signal from unlabelled data. 
In general, the form of $p(\theta)$ may be unknown or aligning the empirical distribution of unlabelled predictions to it may be non-trivial. However, in classification tasks where  $\smash{y|x}$ is \textit{deterministic}, i.e.~each $\smash{x\!\in\!\mathcal{X}}$ has a unique label, $\smash{p(\theta)}$ simplifies to a \textit{discrete} distribution and aligning the distribution of unlabelled predictions to it can be achieved by standard gradient-based optimisation methods by approximating the discrete $\smash{p(\theta)}$ with a suitable \textit{continuous relaxation} $\smash{q(\theta)}$.  

Stepping back, it may seem counter-intuitive to tackle SSL with discriminative methods that rely on $p(y|x)$, rather than those pertaining to $p(x)$, when fewer labels are available by its definition. 
However, the latter methods require additional knowledge  of $p(x)$, e.g.~domain-specific invariance, which may not always be available; and, where it is, the two approaches can be successfully combined, as in recent state-of-the-art methods \citep{berthelot2019mixmatch, berthelot2019remixmatch}, making it relevant to understand discriminative approaches.

The key contributions of this work are: 
\vspace{-8pt}
\begin{itemize}[leftmargin=0.5cm]    
    \item to propose a probabilistic model for \textit{discriminative} SSL (DSSL), comparable to the classical generative model, contributing to the theoretical understanding of semi-supervised learning (\sect{\ref{sec:DSSL}});
    \vspace{-2pt}
    \item 
    to justify several previous SSL methods, e.g.~\textit{entropy minimisation}, as DSSL under the assumption $y|x$ is \textit{deterministic}, and to propose a new \textit{deterministic prior} loss that improves upon them;
    and
    \vspace{-2pt}
    \item to show that the DSSL model extends also to a family of (often distinct) neuro-symbolic SSL methods, to rigorously justify and unify them with `regular' SSL (\sect{\ref{sec:NSL}}), contributing to bridging the gap between \textit{connectionist} and  \textit{symbolic} approaches. 
\end{itemize}

\vspace{-5pt}
\section{Background and related work}
\label{sec:bg}
\vspace{-5pt}

Notation:
$\smash{\mX\! \!=\! \{x^i\}_{i=1}^{n}}$, $\smash{\mY\! \!=\! \{y^i\}_{i=1}^{n}}$ are labelled data, treated as samples of random variables $\rx$, $\ry$, with domains $\mathcal{X}, \mathcal{Y}$;
$\smash{\mX'\! \!=\! \{x^j\}_{j=1}^{m}}$, $\smash{\mY'\! \!=\! \{y^j\}_{j=1}^{m}}$ are unlabelled data and their unobserved labels. 
Each $\theta$ parameterises a distribution $p(y|x)$ and is treated as a realisation of random variable $\uptheta$ in domain $\Theta$.
$\theta_k$ denotes component $k$ of $\theta$.
(Subscripts are dropped where possible to lighten notation.)

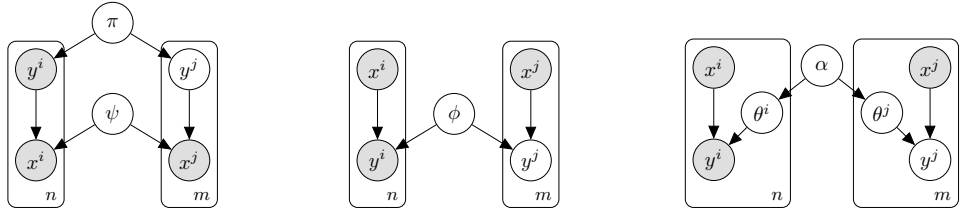
\begin{figure*}[!tb]
    \centering
    \begin{subfigure}[b]{0.3\textwidth}
        \centering{
        \resizebox{3.0cm}{!}{

\begin{tikzpicture}[scale=1., every node/.style={transform shape}]


  \node[obs]                                    (x)     {$x^i$};
  \node[obs,    above=.85cm of x]               (y)     {$y^i$};
  \node[latent, right=.6cm of x, yshift=.8cm]   (t)     {$\psi$};
  \node[latent, right=.6cm of y, yshift=.8cm]   (p)     {$\pi$};
  \node[obs,    right=.6cm of t, yshift=-.8cm]  (xu)    {$x^j$};
  \node[latent, above=.85of xu]                 (yu)    {$y^j$};

  \edge {y}  {x} ; %
  \edge {yu} {xu} ; %
  \edge {t}  {x} ; %
  \edge {t}  {xu} ; %
  \edge {p}  {y} ; %
  \edge {p}  {yu} ; %

  \plate {yx}  {(y)(x)} {$n$} ;
  \plate {yxu} {(yu)(xu)} {$m$} ;

\end{tikzpicture}

        }}
    \end{subfigure}
    \hfill
    \begin{subfigure}[b]{0.3\textwidth}
        \centering
        \resizebox{3.0cm}{!}{

\begin{tikzpicture}[scale=1., every node/.style={transform shape}]


  \node[obs]                                    (y)     {$y^i$};
  \node[obs,    above=.85 of y]                 (x)     {$x^i$};
  \node[latent, right=.6cm of y, yshift=.8cm]   (t)     {$\phi$};
  \node[latent, right=.6cm of t, yshift=-.8cm]  (yu)    {$y^j$};
  \node[obs,    above=.85 of yu]                (xu)    {$x^j$};

  \edge {x}  {y} ; %
  \edge {xu} {yu} ; %
  \edge {t}  {y} ; %
  \edge {t}  {yu} ; %

  \plate {yx}  {(x)(y)} {$n$} ;
  \plate {yxu} {(xu)(yu)} {$m$} ;

\end{tikzpicture}

        }
    \end{subfigure}
    \hfill
    \begin{subfigure}[b]{0.35\textwidth}
        \centering
        \resizebox{3.85cm}{!}{

\begin{tikzpicture}[scale=1., every node/.style={transform shape}]

  \node[latent]                                      (a)     {$\alpha$};
  \node[latent, left =.3cm of a, yshift= -.8cm]      (t)     {$\theta^i$};
  \node[obs,    left =.1cm of t, yshift= -.8cm]      (y)     {$y^i$};
  \node[obs,    above=.85cm of y]                    (x)     {$x^i$};
  \node[latent, right=.3cm of a, yshift=-.8cm]       (tu)    {$\theta^j$};
  \node[latent, right=.1cm of tu, yshift=-.8cm]      (yu)    {$y^j$};
  \node[obs,    above=.85cm of yu]                   (xu)    {$x^j$};

  \edge {x} {y} ; %
  \edge {t} {y} ; %
  \edge {xu} {yu} ; %
  \edge {tu} {yu} ; %
  \edge {a} {t} ; %
  \edge {a} {tu} ; %

  \plate {yx} {(x)(y)(t)} {$n$} ;
  \plate {yxu} {(xu)(yu)(tu)} {$m$} ;




\end{tikzpicture}
        }
    \end{subfigure}
    \caption{Graphical models for: generative SSL \figleft; discriminative SSL (previous \citep{chapelle2009semi}) \figcenter; discriminative SSL (ours) 
    \figright. Shading indicates variables are \textit{observed} (else \textit{latent}). 
    }
    \label{fig:graph_models}
    \vspace{-18pt}
\end{figure*}

\textbf{Semi-supervised learning (SSL)}
%
is a well-established field, covered by several surveys and taxonomies \citep{seeger2006taxonomy, zhu2009introduction, chapelle2009semi, van2020survey}. 
Methods can be categorised by
how they adapt supervised learning algorithms \citep{van2020survey}; or 
their  assumptions \citep{chapelle2009semi}, such as that the data of each class form a cluster/manifold, or that different classes are separated by low density regions. It has been suggested that all such assumptions are variations of \textit{clustering} \citep{van2020survey}. Although clustering is not well defined \citep{estivill2002so}, from a probabilistic perspective this suggests that SSL methods assume $p(x)$ to be a \textit{mixture} of class-conditional distributions that are distinguishable by some property, satisfying the condition that for unlabelled $x$ to help in learning to predict $y$ from $x$, the distribution of $x$ must contain information relevant to the prediction \citep{chapelle2009semi, zhu2009introduction}. We categorise SSL methods according to the properties of $p(x)$ they leverage.

A canonical SSL method that relies on explicit assumptions of $p(x)$ is the classical generative model:
\vspace{-5pt}
\begin{align}
    p(\mX\!, \mY\!, \mX')     
    = \int_{\psi, \pi} p(\psi, \pi) p(\mX|\mY\!, \psi)p(\mY | \pi) \,
        {\sum}_{\mY^{'}}p(\mX'| \mY'\!, \psi)p(\mY' |\pi)
    \label{eq:ssl_generative}
\end{align}

\vspace{-12pt}
Parameters $\psi,\pi$ of $p(x|y)$ and $p(y)$ are learned from labelled and unlabelled data (e.g.~via the EM algorithm), and predictions $p(y|x) \!=\! p(x|y)p(y)/p(x)$ follow by Bayes' rule. 
Fig.~\ref{fig:graph_models} \figleft\ shows the corresponding graphical model.
Whilst generative SSL has an appealing probabilistic rationale, it is rarely used in practice, similarly to its supervised counterpart, because $p(x|y)$ is often complex yet must be accurately modelled \citep{grandvalet2005semi, zhu2009introduction, lawrence2006gaussian}. 
That said, domain-specific invariances may be known without knowing $p(x|y)$ in full, e.g.~translation-invariance in images, allowing \textit{data augmentation} and \textit{consistency regularisation} methods \citep{sajjadi2016regularization, laine2017temporal,tarvainen2017mean, miyato2018virtual} that adapt real $x$ samples into artificial samples expected to be of the same class, even if that is unknown.
Other SSL methods consider $p(x)$ in terms of components $p(x|z)$, where $z$ is a latent representation useful for predicting $y$ \citep{kingma2014semi, rasmus2015semi}.

The SSL methods on which we focus take a particular \textit{discriminative} approach: a parametric function $f_\omega\!:\mathcal{X}\!\to\!\Theta$ (typically a neural network) predicts $\theta$ as a function of $x$, $\smash{f_\omega(x)\!\doteq\!\tilde\theta\!\approx\!\theta}$; and 
 a function of unlabelled predictions $\smash{\ell^{\textit{\,u}} \!=\! \sum_j l(\tilde\theta^j)}$ is added to a negative log-likelihood loss function. 
Such methods are often applied to $K$-class classification where $\theta$ is a vector on the simplex $\Delta^K\!\subset\!\sR^K\!$ and $p(y|x)$ is multinomial.
\textbf{Entropy minimisation} \citep{grandvalet2005semi} assumes classes are ``well separated'' and uses entropy of $p(y|x)$ as a proxy for class overlap.
\textbf{Mutual exclusivity} \citep{sajjadi2016mutual, xu2018semantic} assumes no class overlap whereby predictions form one-hot vectors that, seen as logical variables $\vz$, satisfy the formula $\smash{\bigvee_k (\vz_k \bigwedge_{j\ne k}\!\neg \vz^j)}$, from which $\ell^{\textit{\,u}}$ is derived.
\textbf{Pseudo-labelling} \citep{lee2013pseudo} treats currently predicted class labels $\smash{k^* \!=\! \argmax_{k} \theta_k}$ for unlabelled data as though true labels.
Table~\ref{tab:implied_p_theta} (col.~1) shows the loss component $\smash{l(\tilde\theta)}$ each method applies to unsupervised data.
Although intuitive, these methods lack theoretical justification comparable to generative SSL (Eq.~\ref{eq:ssl_generative}). 
In this respect, \citep{lawrence2006gaussian} notes that summing over all labels for unlabelled data under the graphical model in Fig.~\ref{fig:graph_models} (\textit{centre}) is of no use:
\vspace{-5pt}
\begin{align}
    p( \mY| \mX, \mX') 
    = \int_\theta p(\phi)  p(\mY |\mX, \phi)  
        \underbrace{\mathsmaller{\sum_{\mY^{'}}} p(\mY'|\mX', \phi)}_{=1}
    = \int_\phi p(\phi)  p(\mY |\mX, \phi).
    \label{eq:no_use}
\end{align}

\vspace{-11pt}
Indeed, parameters $\phi$ of $p(\mY |\mX\!, \phi)$ are provably independent of $\mX'$ \citep{seeger2006taxonomy, chapelle2009semi}. 
To break the independence, previous works introduce additional variables \citep{lawrence2006gaussian}, or assume that parameters of $p(y|x)$ are dependent on those of $p(x)$ \citep{seeger2006taxonomy}. We extend this line of research to propose a hierarchical latent variable model for \textit{discriminative} SSL (DSSL), analogous to that for generative SSL (Eq.~\ref{eq:ssl_generative}).

\begin{figure}[t!]
    \centering
    \resizebox{11.7cm}{!}{
    \input{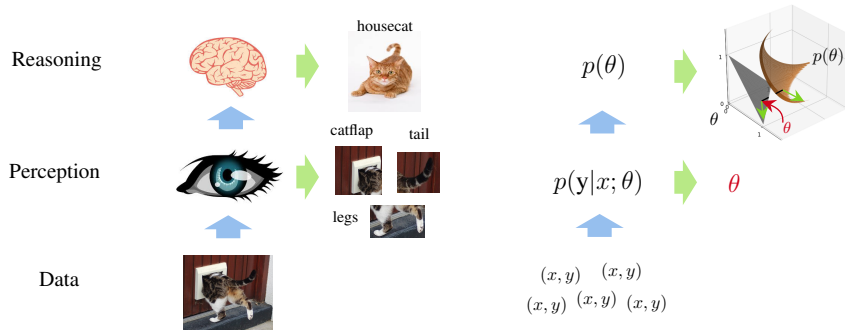}
    }
    \vspace{-2pt}
    \caption{An NSL  framework combining statistical learning (perception) and logical rules (reasoning) \citep{valiant2000neuroidal, garcez2019neural}. 
    Comparison is drawn to the DSSL model (\sect{\ref{sec:DSSL}}), in which logical rules may define $p(\theta)$ (\sect{\ref{sec:NSL}}).
    }
    \label{fig:NSL_framework}
    \vspace{-16pt}
\end{figure}
%

\textbf{Neuro-symbolic learning (NSL)}
%
combines statistical machine learning, often using neural networks, and logical reasoning \cite[e.g.~see][]{garcez2019neural}. Approaches often introduce statistical methods into a logical framework \citep[e.g.][]{rocktaschel2017end, manhaeve2018deepproblog}; or inject logical rules into statistical learning methods \citep[][]{rocktaschel2015injecting, ding2018improving, marra2019integrating, van2019semi, wang2019satnet}. 
Figure~\ref{fig:NSL_framework} shows a conceptual framework for NSL  \citep[][]{valiant2000neuroidal, garcez2019neural} that places statistical methods within a low-level \textit{perceptual} component that processes raw data (e.g.~performing pattern recognition), which feeds a \textit{reasoning} module, e.g.~performing logical inference. This template can be seen in many NSL works \citep[e.g.][]{wang2019satnet, dai2019bridging}; those closest to our own propose a 2-layer graphical model comprising a neural network and a ``semantic layer'' \cite[Fig.~1]{marra2019integrating}, and a graphical model for SSL comprising a neural network component and a logic-based prior \cite[Fig.~1]{van2019semi}. 
By comparison, where \cite{marra2019integrating} introduces logical constraints as a design choice (their Eq.~2), in our DSSL model, logical rules innately define the support of a probability distribution. In \cite{van2019semi}, knowledge base rules directly influence \textit{labels} of only \textit{unlabelled} data, whereas under the DSSL model such rules affect \textit{parameters} of \textit{all} label distributions $p(y|x; \theta)$. 
At an intuitive level, where \cite{van2019semi} treats probabilities as ``continuous relaxations'' of logical rules, the DSSL model treats logical rules akin to limiting (discrete) cases of continuous probability distributions.
We note that many other works consider comparable latent variable models (e.g. treating logical rules as constraints in a quasi-variational Bayesian approach \cite{mei2014robust}) or structured label spaces \cite[e.g.~see][]{zhu2009introduction}, but we restrict attention to neuro-symbolic approaches for SSL.

\begin{figure}[t!]
    \parbox{.5\linewidth}{
        \captionof{table}{\small Per-sample unsupervised loss component $\smash{l(\tilde\theta)}$ and implied $\smash{q(\tilde\theta)}$ (up to prop.)  for  DSSL methods: 
            minimum entropy (E), 
            mutual exclusivity (X), 
            pseudo-label (PL), and 
            deterministic prior (DP, ours, see \sect{\ref{sec:scenarios}}). 
        }
        \label{tab:implied_p_theta}        \resizebox{7.3cm}{!}{
        \hspace{-8pt}
        \renewcommand{\arraystretch}{1.}
        \begin{tabular}{c  l l}
            \toprule
            & $l(\tilde\theta) \quad [\ell^u\!\doteq\!\sum_jl(\tilde\theta^j)]$  &  $\propto q(\tilde\theta)$\\
            \toprule 
            \textbf{E}      & $\sum_{k} \tilde\theta_k\log \tilde\theta_k$ 
                        &  $\prod_k {\tilde\theta}_k^{\,\tilde\theta_k}$ \\
            \textbf{X}      & $ \log\sum_{k\!} \tilde\theta_k
                            \prod_{k'\ne k}1 \!-\! \tilde\theta_{k'}$
                &  $\sum_k \tilde\theta_k  \prod_{k'\ne k}1 \!-\! \tilde\theta_{k'}$\\
            \textbf{PL}      & $\log \sum_k \mathbbm{1}_{k=k^*}\tilde\theta_k\qquad\ $               
                        &  $\max_k \tilde\theta_k$ \\
            \textbf{DP}     & $\log\sum_{k} \tilde\theta_k^{\,T}$ 
                        &  $\sum_k {\tilde\theta_k^{\,T}}$ \\
            \bottomrule
        \end{tabular}
        }

    }
    \hfill
    \parbox{.49\linewidth}{
        \centering
        \includegraphics[trim={5 5 15 5}, clip, width=.43\textwidth]{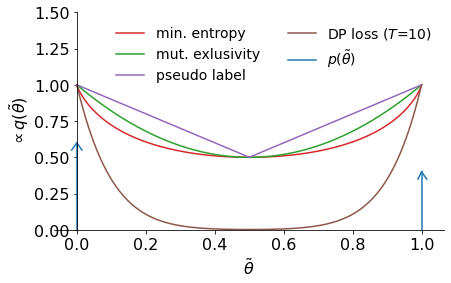}
        \vspace{-7pt}
        \captionof{figure}{Relaxations $q(\tilde\theta)$ of the DSSL prior.
    }
    \label{fig:p_theta_priors}
    }
    \vspace{-15pt}
\end{figure}

\vspace{-5pt}
\section{Probabilistic model for discriminative semi-supervised learning}
\label{sec:DSSL}
\vspace{-5pt}

Here we present the probabilistic model for discriminative semi-supervised learning methods (DSSL) in which a parametric function $f_\omega$ (with \text{weights} $\omega$) learns to map each data point $x\!\in\!\mathcal{X}$ to parameter $\theta\!\in\!\Theta$ of the respective distribution $p(y|x)$. Our running example is $K$-class classification, where $\theta$ is the mean parameter of a multinomial label distribution and its domain $\Theta$ is the simplex $\Delta^K\!\!\subset\!\mathbb{R}^K\!$. For clarity, we emphasise that $f_\omega$ maps $x$ to a label distribution \textit{parameter}, not a particular label $y$. Whilst, in principle, different $x$ could have arbitrarily different label distributions, it is implicitly assumed that similar $x$ have (somewhat) similar label distributions by choosing $f_\omega$ to be continuous, and also sufficiently flexible to approximate the ground truth $f(x)\!=\!\theta$ (e.g.~a neural network). 

The proposed model treats parameters $\theta$ as latent random variables with distribution $p(\theta)$. Figure~\ref{fig:graph_models} (\textit{right}) shows the corresponding graphical model with $p(\theta)$ parameterised by $\alpha$. Omitting $\alpha$ for brevity and letting
 $\theta \!=\! \{\theta^{\mathsmaller{\mX\!\!}}, \theta^{\mathsmaller{\mX'\!}}\}$, 
 $\theta^\mathsmaller{\mX}\!\!=\!\{\theta^i\}_{i=1}^n, \theta^{\mathsmaller{\mX'}}\!\!\!=\!\{\theta^j\}_{j=1}^m$, 
 the conditional likelihood is given by:
\vspace{-5pt}
\begin{align}
    p( \mY| \mX\!, \mX') 
    &= \!
    \int_{\theta} p(\mY | \mX,\theta^\mathsmaller{\mX}) p(\theta^\mathsmaller{\mX})  
        \,\underbrace{{\mathsmaller{\sum_{\mY^{'}}}} 
        p(\mY'| \mX'\!,\theta^{\mathsmaller{\mX'}})}_{=1}  p(\theta^{\mathsmaller{\mX'}})
    \nonumber\\
    &= \!
    \int_{\theta} \prod_i p(y^i | x^i, \theta^i) p(\theta^i)  
        \prod_j p(\theta^j)
    \qquad \overset{\dagger}{\approx} \quad 
    \prod_j p(y^i| x^i, \theta^{i*})  p(\theta^{i*}) \!
    \int_{\theta} \!\prod_i p(\theta^j)  
    \label{eq:SSL_new}
\end{align}

\vspace{-11pt}

The approximation ($\dagger$) assumes that parameters of \textit{labelled} data $\theta^i$ are learned with sufficient certainty that posterior distributions $p(\theta^i|x^i,y^i) \!\propto\! p(y^i|x^i,\theta^i)p(\theta^i)$ 
are well approximated by delta functions $\smash{\delta_{\theta^i-\theta^{i*}}}$ (discussed further in \sect{\ref{sec:scenarios}}).
Rather than considering all possible parameter values, $\theta$ is substituted by $\smash{\tilde\theta\!\doteq\!f_\omega(x)}$ and a \textit{maximum a posteriori} approach taken by maximising (w.r.t. $\omega$): 
\vspace{-5pt}
\begin{align}
    \ell_{\text{DSSL}}(\omega)
    =
     \sum_i\sum_{k} y^i_k\log \tilde\theta^i_k  
    + \sum_i \log p(\tilde\theta^i|\alpha) 
    + \sum_j \log p(\tilde\theta^j|\alpha) 
    \label{eq:DSSL}
\end{align}

\vspace{-11pt}
Here, the first term encourages $\smash{\tilde\theta^i\!\approx\theta^{i*}}$, so that the model learns the desired parameter for labelled data, as in supervised learning. In principle, the middle term allows parameters $\alpha$ of $p(\theta)$ to be learned from the labelled data. In the last term, parameter predictions for unlabelled data $\smash{\tilde\theta^j}\!$, as influenced by the predictions of nearby labelled data (due to continuity of $f_\omega$), are encouraged towards a local mode of $p(\theta)$.\footnote{
    To put this another way, note that applying $f_\omega$ to $x\!\sim\!p(x)$ induces an empirical distribution over predictions $\smash{p_\omega(\tilde\theta)}$,  dependent on $\omega$. The last term effectively minimises the KL divergence 
    $D_{\mathsmaller{\!K\!L}}[p_\omega(\uptheta)||p(\theta|\alpha)] \!\doteq\! 
    \int_\theta p_\omega(\theta)\log \tfrac{p_\omega(\theta)}{p(\theta|\alpha)}$, but \textit{ignoring} the entropy of $p_\omega(\uptheta)$ that would prevent $p_\omega(\uptheta)$ `collapsing' to modes of $p(\theta)$.}
In general, the analytic form of $p(\theta)$ required in Eq.~\ref{eq:DSSL} may not be known, or encouraging predictions to the modes of $p(\theta)$ could be undesirable, however, in the cases we are interested in where $y|x$ is \textit{deterministic}, we see in \sect{\ref{sec:scenarios}} that both concerns are satisfied.

We briefly highlight the symmetry between the two probabilistic models for SSL, slightly restating Eq.~\ref{eq:ssl_generative} for clearer comparison (e.g.~omitting $\pi$ for brevity) together with the joint equivalent of Eq.~\ref{eq:SSL_new}:
\vspace{-5pt}
\label{sec:relationship}
\begin{align}
    p(\mX\!, \mY\!, \mX')     
    &= \int_{\bm{\psi}}
        p(\bm{\psi}) p(\mX|\mY\!, \bm{\psi}^{\mathsmaller{\mY}})p(\mY) \,
        {\sum}_{\mY^{'}}p(\mX'| \mY'\!, \bm{\psi}^{\mathsmaller{\mY'}})p(\mY')    
    \tag*{(G)}\nonumber
    \\
    &= \int_{\theta} \ 
        p(\theta) \  p(\mY|\mX\!, \theta^{\mathsmaller{\mX}})p(\mX) \,
        {\sum}_{\mY^{'}}p(\mY'| \mX'\!, \theta^{\mathsmaller{\mX'}})p(\mX')    
    \tag*{(D)}\nonumber
\end{align}

\vspace{-10pt}
Under the generative model, a conditional distribution parameter $\psi$ is sampled and assigned to (or \textit{indexed} by) each value in the domain $\mathcal{Y}$ (i.e.~each label); as $y$ are then sampled, their corresponding parameter $\psi$ (a latent variable) defines a distribution from which $x$ is sampled. Parameters $\psi$ are learned for each class $k$, equivalent to an \textit{implicit} \text{mapping} $f(k)\!=\!\psi_k$.
The discriminative model follows analogously: parameters $\theta$ are notionally sampled and assigned to every value in $\mathcal{X}$; as $x$ are sampled, their corresponding parameter $\theta$ defines the distribution from which a label $y$ is drawn. Here, the mapping $f(x)\!=\!\theta$ is learned \textit{explicitly}.
Both models can be seen to leverage a distribution across data samples to enable SSL: $p(x)$ in the generative case, $p(\theta)$ in the discriminative case.

\begin{figure*}[!t]
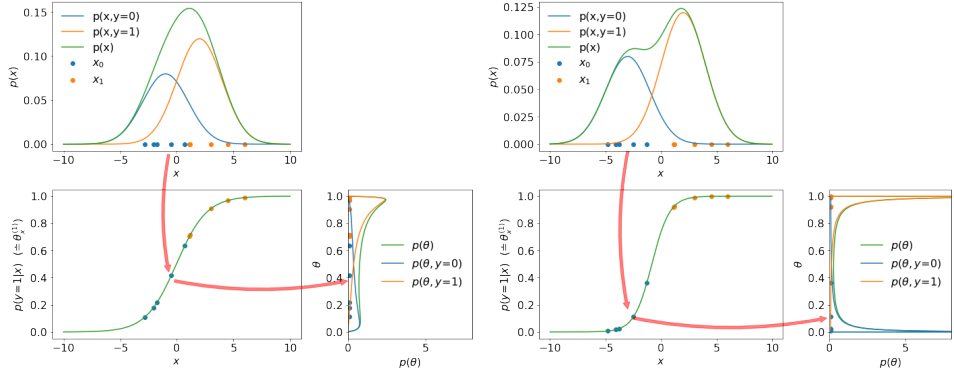

    \centering
    \begin{subfigure}[b]{0.45\textwidth}
        \centering
        \includegraphics[width=\linewidth]{figures/p_theta5_1.pdf}
    \end{subfigure}
    \begin{subfigure}[b]{0.45\textwidth}
        \centering
        \includegraphics[width=\linewidth]{figures/p_theta5_2.pdf}
    \end{subfigure}
    \vspace{-7pt}
    \caption{Probability distributions for a mix of 2 uni-variate Gaussians with different separations of class means $\mu_k$: (top panels) class distributions $p(x|\ry\!=\!k)$; (lower left) the parameter $\theta\!\doteq\!p(y\!=\!1|x)$ corresponding to each $x$; and (lower right) the resulting distribution $p(\theta)$ and components $p(\theta,\ry\!=\!k)$.
    }
    \label{fig:p_theta_mog}
    \vspace{-11pt}
\end{figure*}

\vspace{-5pt}
\section{Applying the discriminative semi-supervised learning model}
\label{sec:scenarios}
\vspace{-5pt}

We now consider implementations of the DSSL model. 
Our main interest is in classification where $y|x$ is \textit{deterministic}, but we first consider a simple \textit{stochastic} scenario to clarify the notion of $p(\theta)$.

$\bullet$ \textbf{Stochastic classification}:
For classification of 2 equivariant 1-D Gaussians, 
$\smash{\rx|y \!=\!k \sim \mathcal{N}(\mu_k, \sigma^2})$ with class probabilities $\pi_k \!=\!p(y\!=\!k)$, 
$p(\theta)$ can be derived in closed form (see Appendix \ref{sec:appx_MOG}).
Fig.~\ref{fig:p_theta_mog} shows $p(x)$ (top panels) and $p(\theta)$ (right panels) for two differences between class means $\mu_k$.
Under the generative SSL model, parameters $\mu_k, \pi_k$ are updated to better explain the unlabelled data $\mX'$. Under DSSL, with no model of $p(x)$, a function learns to approximate $f(x)\!=\!\theta$ (lower left panels) to fit the labelled data \textit{and} so that the distribution of unlabelled predictions reflects $p(\theta)$. Although both SSL models can be used here, the analytical form of $p(x)$ is typically unknown or too complex to model, whereas a good approximation to $p(\theta)$ may be both known and far simpler.

In contrast, in many tasks, each $x$ occurs \textit{exclusively} with one label $y$, e.g.~in the MNIST dataset, a particular image of a two is only labelled ``2''. The same is true more generally when the very purpose of labels is to distinguish one item from another. Where so, $y|x$ is \textbf{deterministic}, which we now assume. We distinguish between whether labels represent distinct classes or sets of binary features.

$\bullet$ \textbf{Deterministic classification (distinct classes)}:
If the label domain $\mathcal{Y}$ is a discrete set of $K$ classes and $y|x$ is deterministic, each distribution $p(y|x)$ equates to an indicator function with parameter $\theta$ at a vertex $\ve_k$ of the simplex $\Delta^{K\!}$, i.e.~all $\theta\!\in\!\{\ve_k\}_{k=1}^K$ are \textit{one-hot}.
With only those values possible, although $p(\theta)$ is defined over the continuous domain $\smash{\Delta^K\!}$, it effectively reduces to a \textit{discrete} distribution given by a sum of delta functions $\sum_k\pi_k\delta_{\theta-\ve_k}$  weighted by class probabilities $\pi_k\!=\!p(y\!=\!k)$.
(This can be seen as a limiting case of the stochastic example where overlap of class conditional distributions is reduced by increasing class mean separation or reducing class variance.) 

For semi-supervised learning, this means that assumption ($\dagger$) in Eq.~\ref{eq:SSL_new} is immediately more plausible since each parameter $\theta$ is fully determined by a single observation $(x,y)$, rather than requiring multiple samples and being subject to sampling error. Also, the analytic form of $p(\theta)$ is available to substitute into Eq.~\ref{eq:DSSL}.
However, this discrete $p(\theta)$ has zero support for any prediction $\smash{\tilde\theta}$ that is not precisely one-hot and provides no gradient to update $\omega$. As such, $p(\theta)$ can be substituted by a suitable \textit{relaxation} $q(\theta)$.
Lastly, since parameters for labelled data are accurately learned from the data, applying the prior is largely redundant and the middle term in Eq.~\ref{eq:DSSL} can be dropped, to give:
\vspace{-3pt}
\begin{align}
    \label{eq:DSSL_det}
    \ell_{\text{det-DSSL}}(\omega)
    =
     \sum_i\sum_{k} y^i_k\log \tilde\theta^i_k  
    + \sum_j \log q(\tilde\theta^j) \ ,
\end{align}

\vspace{-9pt}

a general loss function for \textit{deterministic} discriminative SSL. 
The last term may be viewed as \textit{regularising} a supervised learning model, but note it is a function of model \textit{outputs} $\smash{\tilde\theta}$ not weights $\omega$, as is common (e.g.~$\ell_1$, $\ell_2$). $q(\theta)$ can also be considered a \textit{critic} of unlabelled predictions, providing a means of updating them (via $\omega$) to be more plausible.
Comparing Eq.~\ref{eq:DSSL_det} to existing methods (\sect{\ref{sec:bg}}), the final term gives a probabilistic rationale for adding a function $\ell^u$ of the unlabelled predictions to a supervised loss function, as seen in \text{entropy minimisation} \citep{grandvalet2005semi}, \text{mutual exclusivity} \citep{sajjadi2016mutual, xu2018semantic} and \text{pseudo-labelling} \citep{lee2013pseudo}. Accordingly, those methods are probabilistically justified and unified as instances of Eq.~\ref{eq:DSSL_det} for choices of $q(\theta)$ (up to proportionality) shown in Table \ref{tab:implied_p_theta} and plotted in Fig.~\ref{fig:p_theta_priors}.  
(In practice, $q(\theta)$ need not be normalised since optimisation depends on \textit{relative} gradients of $q(\theta)$.)

\textit{\textbf{Choosing $\bm{q(\theta)}$}}:
The DSSL model does not justify one choice of $q(\theta)$ over another, beyond a need to approximate $p(\theta)$. However, some prior methods may appear to have other theoretical justification, e.g.~minimising \text{entropy} \cite{grandvalet2005semi} or satisfying various axioms \cite{xu2018semantic}. Fig.~\ref{fig:p_theta_priors} shows that the $q(\theta)$ of prior methods are locally maximal at simplex vertices, but do not otherwise closely approximate $p(\theta)$. 

\vspace{-1pt}
Intuitively, the general DSSL approach can be seen to leverage what $f_\omega$ learns from labelled data to make \textit{proto-predictions} for unlabelled data that are better than random; hence updating $f_\omega$ to move them nearer to simplex vertices, where true predictions reside, improves the prediction model on average. The gradient $\smash{q'(\theta)\!=\!\tfrac{dq}{d\theta}}$ determines which proto-predictions have greatest effect in updating $f_\omega$.
It therefore seems appropriate to choose $q(\theta)$ such that the better a proto-prediction resembles a true prediction (i.e.~the nearer to a simplex vertex) the more it influences the update of $f_\omega$ (the higher $q'(\theta)$). Conversely, `uncertain' proto-predictions far from simplex vertices should have little effect.

\textit{\textbf{Deterministic prior (DP)}}: Following this intuition, we construct a new relaxation to $p(\theta)$ by replacing each $\delta_{\theta-\ve_k}$ term
by $\theta_k^{\,T\!}$, a `spike' at $\ve_k$ parameterised by $T$, similar to \textit{temperature} \cite{Goodfellow-et-al-2016, berthelot2019mixmatch} (see Table~\ref{tab:implied_p_theta}; Fig.~\ref{fig:p_theta_priors}). Note, $q_{\text{\tiny DP}}(\theta)\!\to\!p(\theta)$ as $T\!\!\to\!\infty$. Our aim is not to find an optimal $q(\theta)$,  but to test the hypothesis that previous $q(\theta)$ are not justified beyond approximating $p(\theta)$, by better approximating $p(\theta)$.
We compare performance of each $q(\theta)$ using architecture (Wide ResNet ``WRN-28-2'' \cite{zagoruyko2016wide}), image datasets (MNIST, SVHN, CIFAR-10) and methodology of previous SSL studies \cite{oliver2018realistic, berthelot2019mixmatch} (see Appendix \ref{sec:experimental_conditions} for implementation details). Results in Table~\ref{tab:results} show that \textit{DP loss} matches or slightly outperforms prior DSSL methods across all datasets considered. (We note that the performance of \textit{DP loss} is broadly insensitive to $T$ across a range of values. $T\!=\!10$ is used for all datasets.)

\begin{table}[t!]
        \centering
        \caption{Test set accuracy (mean $\pm$std err over 10 runs) for DSSL methods. ($\#$) = num labels $(n)$.}
        \label{tab:results}
        \vspace{1pt}
        \resizebox{11.0cm}{!}{
        \begin{tabular}{l  r l  r l  r l}
            \toprule
            Model &             \multicolumn{2}{c}{MNIST (100)} &\multicolumn{2}{c}{SVHN (1000)} & \multicolumn{2}{c}{CIFAR-10 (4000)} \\
            \toprule 
            Fully supervised (all data: $\mX\!\cup\!\mX'$)   &99.50 & $\pm$0.01       & 97.02 & $\pm$0.05         & 94.63 & $\pm$0.06 \\
            \midrule 
            Deterministic Prior, DP ($T\!=\!10$) &\textbf{97.07}&$\pm$0.19 &\textbf{91.32}&$\pm$0.12 &\textbf{84.86}&$\pm$0.14 \\
            Minimum entropy \cite{grandvalet2005semi}         & \textbf{97.06}&$\pm$0.19       & 90.63 & $\pm$0.15      & 84.57 & $\pm$0.08 \\
            Mutual Exclusivity \cite{sajjadi2016mutual, xu2018semantic}     & 96.58 & $\pm$0.18       & 90.36 & $\pm$0.21        & 84.37 & $\pm$0.09 \\
            \midrule 
            Supervised ($\mX$ only) & 90.99 & $\pm$0.59       & 86.11 & $\pm$0.23        & 82.58 & $\pm$0.06 \\
            \bottomrule
        \end{tabular}
        }
    \vspace{-17pt}
\end{table}

\begin{figure}[b!]
    \vspace{-7pt}
        \centering
        \resizebox{13cm}{5cm}{
        \includegraphics[trim={20 25 50 30}, width=\textwidth]{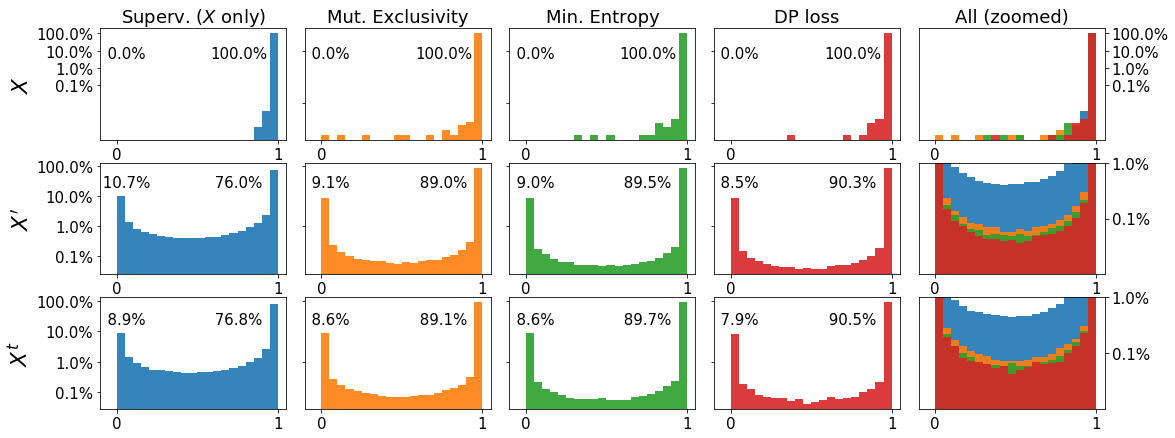}
        }
        \caption{Histograms of predictions $\tilde{\theta}^{k^*}\!\!$ assigned to the true class $k^*$ ($\tilde{\theta}^{k^*}\!\!\!=\!1\Rightarrow\,\,$correct) for all (SVHN) data, split by training 
            (labelled $\mX$, 
            unlabelled $\mX'$) and 
            test $(\mX^{\text{t}})$ set.
        y-axes on log scale.}
        \label{fig:hists}
\end{figure}

\vspace{-1pt}
To analyse whether the choice of $q(\theta)$ has the effect intuited above, Fig.~\ref{fig:hists} shows histograms of the prediction $\smash{\tilde{\theta}^{k^*}\!}$ assigned to each true class $\smash{k^*}$, which should always be 1, for all SVHN data, split by training (labelled and unlabelled) and test set. As expected, all models do well on the labelled training data (top row) and the distribution of learned parameters suggests that $y|x$ is indeed deterministic. All models make errors on unlabelled and test data (low predictions), but the DSSL methods encourage predictions towards simplex vertices (0 or 1), making fewer in between  (see overlay, bottom right). Fewer intermediate predictions can be seen to correlate with performance (Table \ref{tab:results}) and the extent to which unlabelled predictions are encouraged to align with $p(\theta)$ by the gradient of $q(\theta)$ (Fig.~\ref{fig:p_theta_priors}).

$\bullet$ \textbf{Deterministic classification (binary features)}:
In some classification tasks, label vectors $\vy \!\in\! \{0,1\}^K$ represent $K$ binary attributes of the data, e.g.~the 
    presence/absence of features in an image, 
    the configuration of a chessboard or 
    the semantic relations that hold between two knowledge graph entities. 
As previously, $p(\vy|x)$ may be \textit{deterministic} (as those examples demonstrate): whenever a particular $x$ is observed, the same set of attributes occur without stochasticity, and each $x$ has exactly one label $\vy$. 
Considering a multinomial distribution over all $2^K$ possible attribute combinations is typically prohibitive and a classifier learns to predict a vector $\theta\!\in\![0,1]^K\!$, where each $\theta_k$ parameterises a conditional feature distribution $p(\vy_k|x)$.
Analogously to the case of distinct classes, the deterministic assumption restricts each component $\theta_k$ to $\{0,1\}$ and so $\theta$ to $\{0,1\}^{K\!}$, the vertices of the unit hypercube
(equivalent to one-hot vectors). Parameters can be seen to uniquely define labels, and vice versa, under a one-to-one (identity) correspondence between labels and $\theta$ in the support of $p(\theta)$. Accordingly, $p(\theta|\vy) \!=\!\delta_{\theta-\vy}$; and
$p(\theta)$, as required for DSSL (Eq.~\ref{eq:DSSL}),
is again a discrete distribution $p(\theta) 
\!=\! 
\smash{\sum_{\vy}}\pi_\vy\,\delta_{\theta - \vy}$ with marginal label probabilities $\pi_\vy\!=\!p(\vy)$. 
As before, a suitable relaxation $q(\theta)$, e.g.~\textit{DP loss}, enables gradient-based SSL by optimising Eq.~\ref{eq:DSSL_det}. 

[Note, the identity mapping between each label $\vy$ and its corresponding $\smash{\theta}$
suggests that $p(\theta)$ could be learned from \textit{unpaired labels} $\vy\!\sim\!p(\vy)$, an alternative SSL scenario that we leave to future work.]

\vspace{-5pt}
\section{Neuro-symbolic semi-supervised learning}
\label{sec:NSL}
\vspace{-5pt}

When classifying multiple binary features (see \sect{\ref{sec:scenarios}}), certain feature combinations may be impossible, e.g.~an animal having legs \textit{and} fins, 
three kings on a chessboard, or 
knowledge graph entities being related by \textit{capital\_of} but not \textit{city\_in}. 
Here, \textit{valid} attribute combinations form a subset $\sV$ of all \textit{feasible} labels $\sF\!\doteq\!\{0,1\}^K\!$, defined by constraints, such as attributes being mutually exclusive,
the rules of the game, or
relationships between relations. 
Such constraints can often be expressed as a set of \textit{logical rules} and
incorporating them in statistical learning is appealing: they often apply globally, in contrast to the uncertain generalisation in statistical models; and they may allow a large set $\sV$ to be defined succinctly. 
Fig.~\ref{fig:NSL_example} (\textit{left, centre}) gives a simple illustration of $\sV$ and $\sF$ for a set of logical rules $\mathcal{R}$. 

Where $p(\vy|x)$ is \textit{deterministic}, the one-to-one correspondence between labels $\vy\!\in\!\{0,1\}^K\!$ and parameters in the support of $p(\theta)$  (see \sect{\ref{sec:scenarios}}), means that \textit{valid} labels correspond to \textit{valid} parameters (we thus let $\sV/\sF$ denote valid/feasible labels \textit{or} parameters).
It follows that $p(\theta)$ is given by:
\vspace{-4pt}
\begin{equation}
    \label{eq:NSL_p_theta}    
    p(\theta)
    \quad =\quad  
    \sum_{\vy\in\sV}
    p(\vy)p{(\theta|\vy)}
    \quad =\quad  
    \sum_{\vy\in\sV}
    \pi_\vy\delta_{\theta-\vy}
    \,,
\end{equation}

\vspace{-10pt}
where $\pi_\vy \!=\! p(\vy)$ and $\theta\!\in\!\Theta\!=\![0,1]^{K\!}$.
Eq.~\ref{eq:NSL_p_theta} shows that if labels are subject to logical rules, those rules define the \textit{support} of $p(\theta)$, the distribution required for DSSL (Eq.~\ref{eq:DSSL}). 
(Note: Eq.~\ref{eq:NSL_p_theta} also holds for any `larger' set $\sV'$, where $\sV\!\subseteq\!\sV'\!\subseteq\! \sF$.)
Thus, logical rules can be integrated into semi-supervised learning if they can be mapped into the mathematical form of Eq.~\ref{eq:NSL_p_theta}.
By dropping $\smash{\pi_\vy}$ terms in Eq.~\ref{eq:NSL_p_theta}, 
the support of $p(\theta)$ can be defined explicitly as 
$s\!:\Theta\!\to\!\{0,\infty\}$,  which factorises: 
\vspace{-4pt}
\begin{equation}
    \label{eq:s_theta}
    s(\theta)  
    \quad =\quad 
    \sum_{\vy\in\sV} \delta_{\theta-\vy}
    \quad =\quad 
    \sum_{\vy\in\sV} \,
    \prod_{k:\vy_k=1} \!\!\delta_{\theta_k-1}
    \prod_{k:\vy_k=0} \!\!\delta_{\theta_k-0}
    \,.
\end{equation}

\vspace{-10pt}
Each term in the summation of Eq.~\ref{eq:s_theta} effectively tests whether the argument $\theta$ matches a \textit{valid} label $\vy\!\in\!\sV$:
$s(\theta)\!=\!\infty$ if $\theta\!\in\!\sV$, $s(\theta)\!=\!0$ otherwise.
When restricted to \textit{feasible} $\theta\!\in\!\sF$ (i.e.~binary vectors), Eq.~\ref{eq:s_theta} mirrors a \textit{logical formula} in propositional logic over \textit{logical variables}  $\vz_k\!\in\!\{\textit{True, False}\}$:
\vspace{-4pt}
\begin{equation}
    \label{eq:NSL_logical_form}
    t(\vz) \ \ =\ \  \bigvee_{\vy\in\sV} \,
    \bigwedge_{k: \vy_k=1}\!\!\vz_k
    \bigwedge_{k: \vy_k=0}\!\!\!\!\!\!\neg
    \vz_k
    \ ,
\end{equation}

\vspace{-10pt}
Here, $t(\vz)$ evaluates to \textit{True} if and only if $\vz$ corresponds to a valid label $\vy\!\in\!\sV$, in the sense that $\vz_k\!=\!\textit{True}$ \textit{iff} $\vy_k\!=\!1$, for all $k$; hence $t$ and $s$ perform analogous tests of validity.

The relationship between Eqs.~\ref{eq:s_theta} and \ref{eq:NSL_logical_form} reflects a correspondence between logical and algebraic formulae familiar in \textit{fuzzy logic} and \text{neuro-symbolic learning} \citep[e.g.][]{bergmann2008introduction,  serafini2016logic, van2019semi}. 
Under specific
mappings of variables and operators, \textit{satisfiability} (SAT) problems, defined by a set of logical rules over logical variables (e.g.~Eq.\ref{eq:NSL_logical_form}), can be transformed into algebraic functions of binary variables that evaluate to a particular value (often 1) if the constraint is satisfied and 0 otherwise. 

Rather than mapping truth values of logical variables to \textit{values} of binary variables, the transformation of Eq.~\ref{eq:NSL_logical_form} to Eq.~\ref{eq:s_theta} requires an analogous mapping from $\vz_k$ to  $\delta$-functions over $\theta_k\!\in\![0,1]$, indicating whether $\theta_k$ is 0 or 1. Specifically, 
    $\vz_k \!\to\! \delta_{\theta_k-1}$,   
    $\neg\vz_k \!\to\! \delta_{\theta_k-0}$
($\vz_k$ is not defined for $\theta_k\!\not\in\!\{0,1\}$). 
An evaluation to \textit{True} (resp.~\textit{False}) in the logic domain corresponds to $\infty$ (resp.~0) in the numeric. 
Under this mapping,
logical operators $\wedge$ ({\scshape AND}) and $\vee$ ({\scshape OR}) are equivalent to `$\times$' and '+', respectively,  e.g.~$\vz_k\!\wedge\vz_l$  evaluates to \textit{True} \textit{iff} $\delta_{\theta_k-1}\!\times\!\delta_{\theta_l-1}\!=\!\infty$. 
This gives a well-defined mapping between Eqs.~\ref{eq:s_theta} and \ref{eq:NSL_logical_form}: any set of logical rules in the form of Eq.~\ref{eq:NSL_logical_form} can be transformed to a sum of delta functions, each corresponding to a valid variable combination (Eq.~\ref{eq:s_theta}); similarly, any function in the form of Eq.~\ref{eq:s_theta}, possibly learned from the data, can be converted to a set of logical rules (Fig.~\ref{fig:NSL_example}, \textit{left} to \textit{centre})

\begin{figure*}[t!]
    \centering
    \resizebox{11.5cm}{!}{
    \input{figures/fig_logic}
    }
    \caption{Illustration of how a set of logical rules between attributes $\mathcal{R}$ define the support of $p(\theta)$. 
    (\textit{Top left}) $\sF\!=\!\{0,1\}^K$, the set of \textit{feasible} values for $\theta$ if $y|x$ is deterministic (\sect{\ref{sec:scenarios}}). 
    (\textit{Bottom left}) $\mathcal{R}$, a set of logical rules between label attributes. 
    (\textit{Centre}) $\sV\!\subseteq\!\sF$, the set of \textit{valid} values for $\theta$ under the rules $\mathcal{R}$, as encoded by the function $s_\mathcal{R}(\theta)$, the support of $p(\theta)$. 
    (\textit{Right})  $q_\mathcal{R}(\theta)$, a relaxation of $s_\mathcal{R}(\theta)$, the gradient of which can `guide' unlabelled predictions towards valid $\theta$ in DSSL.
    }
    \label{fig:NSL_example}
    \vspace{-17pt}
\end{figure*}

Importantly, this mapping generalises to an arbitrary set of logical rules since  Eq.~\ref{eq:NSL_logical_form} is in \textit{disjunctive normal form} (DNF), a disjunction ($\vee$)  of conjunctions ($\wedge$), and it is well known that any set of logical rules can be written in DNF \cite[][p.102-104]{davey2002introduction}.
    (Note, however, in the worst case, a DNF may involve an exponential number of terms and logical techniques may be required to convert as efficiently as possible, e.g.~as used in \cite{xu2018semantic}.)
Thus, a set of logical rules $\mathcal{R}$ that define valid labels $\sV$, can be written in the form of Eq.~\ref{eq:NSL_logical_form} and so mapped, as above, to a function $s_{\mathsmaller{\mathcal{R}}}(\theta)$ in the form of Eq.~\ref{eq:s_theta}.
This links $\mathcal{R}$ to the analytical form of $p(\theta)$, and so connects logical rules to discriminative SSL (Eq.~\ref{eq:DSSL}). 
Although Eq.~\ref{eq:DSSL} requires $p(\theta)$, logical rules only determine $p(\theta)$ up to probability weights $\pi_\vy$, i.e.~$s_{\mathsmaller{\mathcal{R}}}(\theta)$. Further, as in all deterministic cases, $p(\theta)$ is discrete and a relaxation is required for gradient-based SSL using Eq.~\ref{eq:DSSL_det}. 
Thus, a relaxation $q_{\mathsmaller{\mathcal{R}}}(\theta)$ of $s_{\mathsmaller{\mathcal{R}}}(\theta)$ is used in place of that of $p(\theta)$, which does not appear to harm performance in practice (discussed in Appendix \ref{sec:discussion}).
As previously, $q_{\mathsmaller{\mathcal{R}}}\!:\Theta\!\to\![0,1]$ can be found by substituting $\delta$-functions in $s_{\mathsmaller{\mathcal{R}}}$ by continuous
$g\!:[0,1]\!\to\![0,1]$, where $g(1)\!=\!1$, $g(0)\!=\!0$, as in \textit{DP loss}, to give a function locally maximal only at $\theta\!\in\!\sV$ (Fig.~\ref{fig:NSL_example}, \textit{right}).
This theoretically justifies a family of SSL methods that include functions representing logical rules applied to unlabelled data predictions, and demonstrates how logical rules can fit naturally in a probabilistic framework. 
Specifically, \text{Semantic Loss} \citep{xu2018semantic} is equivalent to choosing $g(\theta_k)\!=\!\theta_k$, 
a common choice in NSL \cite[e.g.][]{serafini2016logic, van2019semi, marra2019integrating}. Previous results (\sect{\ref{sec:scenarios}}) suggest that \textit{DP loss} may provide a good choice for $g$. 
As noted previously (\sect{\ref{sec:DSSL}}), $p(\theta)$ can also be \textit{learned from labelled data} under Eq.~\ref{eq:DSSL}. Now knowing that $p(\theta)$ encodes logical rules over attributes, the DSSL model may also explain approaches that extract rules consistent with observed labels \citep[e.g.][]{wang2019satnet, dai2019bridging}.

\vspace{-5pt}
\section{Conclusion}
\vspace{-5pt}

We present a probabilistic model for \textit{discriminative} semi-supervised learning, analogous to the classical model for generative semi-supervised learning. Central to the DSSL model are parameters $\theta$ of distributions $p(y|x)$, e.g.~as predicted by a typical classifier. Treating those parameters as latent random variables, their distribution $p(\theta)$  serves as a prior over model outputs for unlabelled data.
Where $y|x$ is \textit{deterministic}, the analytical form of $p(\theta)$ is known and \textit{discrete}, enabling the DSSL model to be used. We show that the SSL methods \textit{entropy minimisation}, \textit{mutual exclusivity} and \textit{pseudo-labelling} are explained by the DSSL model for different choices of $q(\theta)$, a relaxation of $p(\theta)$; and that a simple alternative, \textit{deterministic prior}, better reflecting $p(\theta)$ outperforms them. 

Where labels represent the presence/absence of multiple attributes, logical relationships between those attributes may rule out certain combinations. We show that a function representing such rules, familiar in fuzzy logic and NSL, corresponds to the support of $p(\theta)$. Thus a family of neuro-symbolic SSL methods that employ functions representing logical rules are justified under the DSSL model and unified with `regular' SSL. This establishes a principled way to combine statistical machine learning and logical reasoning for semi-supervised learning, fitting a conceptual framework for neuro-symbolic computation \citep{valiant2000neuroidal, garcez2019neural}. 
Possible extensions of this work may combine logical rules with fully supervised learning (Eq.~\ref{eq:DSSL}), or consider SSL with extra labels $y$ rather than $x$ (\sect{\ref{sec:scenarios}}).

\newpage
\bibliographystyle{plainnat}
\bibliography{neurips_2021}

\appendix

\clearpage
\section{Derivation of $p(\theta)$ for Classification of Gaussians}
\label{sec:appx_MOG}
For a general mixture distribution:
%
\begin{align*}
    \theta_k 
    &\ =\  p(y \!=\! k|x) 
    \ =\  \sigma\Big(\log \frac{p(x| y \!=\! k)\pi_k}{\sum_{k' \ne k}p(x| y \!=\! k')\pi_{k'}}\Big);
    \label{eq:theta_x}
\\
    \frac{d \theta_k}{d x} 
    &\ =\  
    \theta_k(1 \!-\! \theta_k)
    \Big(\tfrac{d}{dx} \log p(x| y \!=\! k)
    \ - 
    \sum_{k' \ne k}\frac{p(x| y \!=\! k')\pi_k}{\sum_{k'\!' \ne k}p(x| y \!=\! k'\!')\pi_{k'\!'}}\tfrac{d}{dx}\log p(x| y \!=\! k')\!\Big)  
\end{align*}
For a mixture of 2 equivariate Gaussians, these become:
\vspace{-2pt}
\begin{align*}
    \theta_1 
        &= \sigmoid\big(
        \log\tfrac{\pi_1}{\pi_0} +
        \tfrac{\mu_1-\mu_0}{\sigma^2}x -
        \tfrac{1}{2}(\tfrac{\mu_1^2}{\sigma^2} - \tfrac{\mu_0^2}{\sigma^2})
        \big)
    ,
    \qquad\qquad
    \frac{d \theta_k}{d x} 
        = \theta_k(1-\theta_k)
        (\tfrac{\mu_1^2}{\sigma^2} - \tfrac{\mu_0^2}{\sigma^2})
        .
\end{align*}
Rearranging the former gives $x$ in terms of $\theta$:
\vspace{-2pt}
\begin{align*}
    x
    = 
    \tfrac{\sigma^2}{\mu_1-\mu_0}\big(
    \log\tfrac{\theta_1}{1-\theta_1} 
    - \log\tfrac{\pi_1}{\pi_0}
    + \tfrac{1}{2}(\tfrac{\mu_1^2}{\sigma^2} - \tfrac{\mu_0^2}{\sigma^2})
    \big)
    .
\end{align*}

Substituting into $p(\theta) \!=\! |\tfrac{dx}{d\theta}|p(x)$ gives:
\vspace{-2pt}
\begin{align*}
    p(\theta) 
    &= 
    \sqrt{\tfrac{\sigma^2}{2\pi}}\tfrac{1}{|\mu_1-\mu_0|\theta_0\theta_1}
    \sum_{k=0}^1\pi_k
    \exp\{
    a (\log\tfrac{\theta_1}{\theta_0})^2
    +b_k\log \tfrac{\theta_1}{\theta_0}
    \,+\,c_k
    \}
\end{align*}
where:
\vspace{-2pt}
\begin{align*}
    a &= \tfrac{-\,\sigma^2}{2(\mu_1 - \mu_0)^2},
    \qquad
    b_k = \tfrac{\mu_k}{\mu_1 - \mu_0} + \tfrac{\sigma^2}{(\mu_1 - \mu_0)^2} (\tfrac{\mu_1^2 - \mu_0^2}{\sigma^2} - \log\tfrac{\pi_1}{\pi_0}),
    \qquad
    c_k = -\tfrac{(\mu_1 - \mu_0)^2b_k^2}{2\sigma^2}.
\end{align*}

\section{Experiment Implementation Details}
\label{sec:experimental_conditions}

Our experiments follow the methodology, including hyperparameter choice, of \cite{oliver2018realistic, berthelot2019mixmatch} and use code provided by \cite{zagoruyko2016wide}.\footnote{\url{https://github.com/szagoruyko/wide-residual-networks/tree/master/pytorch}} We run all models over 10 random seeds and report mean and standard error.

\section{Omission of mixture probabilities in the relaxation of $p(\theta)$}
\label{sec:discussion}

In \sect{\ref{sec:NSL}}, we consider relaxations of $p(\theta)$ that restrict attention to the support of $p(\theta)$, i.e.~the discrete locations $\sV\!\subset\!\Theta$ where $p(\theta)$ may be non-zero, and ignore the relative probabilities at each support, given by class probabilities $p(\vy)\!=\!\pi_\vy$. We note that previous discriminative SSL methods ignore class weights also (see Table \ref{tab:implied_p_theta}).
Practical reasons for this are (i) that $\pi_\vy$ may be unknown, and (ii) that unless attributes are independent, i.e.~$p(\vy)\!=\!\prod_kp(\vy_k)$, class probabilities cannot be factorised equivalently to the support, as in Eq.~\ref{eq:s_theta}. 
This is not a theoretical justification  for omitting $\pi_\vy$ terms, hence we consider the validity and possible (non-rigorous) rationale for doing so.

\textbf{Validity}: Considering only the support of $p(\theta)$ is equivalent to assuming a uniform label distribution over that support. Where classes are well-balanced, omitting $\pi_\vy$ is clearly justified, elsewhere to do so might be seen as using a ``partially-uninformative'' prior. 

\textbf{Rationale}: If predictions for unlabelled data were chosen simply to maximise $p(\theta)$, the most commonly occurring label (i.e.~the global mode of $p(\theta)$) would be assigned to all unlabelled data. However, $p(\theta)$ acts on predictions $\smash{\Tilde\theta}$ given by a model that learns to \textit{take class weighting into account}. Thus, where $f_\omega$ predicts a less frequent class for a particular unlabelled data point, intuitively, that signal should be taken into account and not blindly over-ridden by a class weighting in $p(\theta)$. In short, omitting class weights may be appropriate under DSSL since $p(\theta)$ acts as a prior over unlabelled predictions that, to some extent, already take class weights into account. We hope to provide a more rigorous argument in future work.

\end{document}